\documentclass[final]{cvpr}

\usepackage{times}
\usepackage{epsfig}
\usepackage{graphicx}
\usepackage{amsmath}
\usepackage{amssymb}

\usepackage[pagebackref=true,breaklinks=true,colorlinks,bookmarks=false]{hyperref}

\def\cvprPaperID{9983} 
\def\confYear{CVPR 2021}

\begin{document}

\title{Identifying tourist destinations from movie scenes using Deep Learning}

\author{Mahendran Narayanan\\
Independent researcher\\
{\tt\small mahendranNNM@gmail.com}
}

\maketitle

\begin{abstract}
   Movies wield significant influence in our lives, playing a pivotal role in the tourism industry of any country. The inclusion of picturesque landscapes, waterfalls, and mountains as backdrops in films serves to enhance the allure of specific scenarios. Recognizing the impact of movies on tourism, this paper introduces a method for identifying tourist destinations featured in films. We propose the development of a deep learning model capable of recognizing these locations during movie viewing. The model is trained on a dataset comprising major tourism destinations worldwide. Through this research, the goal is to enable viewers to identify the real-world locations depicted in movie scenes, offering a novel way to connect cinema with global travel experiences.
\end{abstract}

\section{Introduction}
Movies possess the remarkable ability to transcend boundaries and languages, leaving a lasting impact on people across the globe. The production teams behind these cinematic creations invest significantly in travel, exploring remote and rarely-visited locations to capture awe-inspiring sceneries. Many of these spots, often transformed into breathtaking cinematic backdrops, become popular tourist destinations, while others remain undiscovered by the world until illuminated through the lens of movies or shared on social media platforms.

However, a notable challenge arises for movie enthusiasts and tourists alike: the lack of specific information about the locations featured in films. Unlike social media platforms where users often share details about the places they visit, movies typically don't provide explicit information about the real-world whereabouts of particular shots. Recognizing this as a drawback, this paper addresses the issue by presenting a novel solution using Deep Learning to identify and recognize the locations depicted in favorite movies.

Traditionally, tourist destinations are equipped with guides who narrate the stories of the places and the movies filmed in and around them. This paper seeks to bridge the gap between movie enthusiasts and travelers by creating a platform that connects people with a shared interest in both cinematic storytelling and the beauty of the landscapes captured in films. The ultimate goal is to enhance the viewing experience for movie buffs who wish to delve deeper into the locations portrayed on screen, fostering a sense of connection between the world of cinema and the diverse landscapes that adorn our planet. Through this innovative approach, the paper aims to provide a valuable resource for those seeking to explore the intersections of film, travel, and cultural discovery.

\begin{figure}
    \centering
    \includegraphics[height=8cm,width=\linewidth]{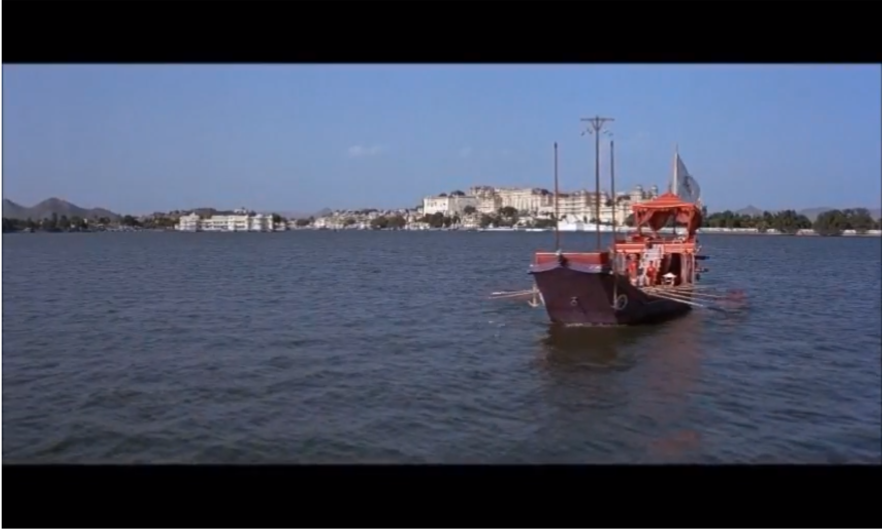}
    \caption{Lake Pichhola}
    \label{fig:lake}
\end{figure}

\begin{figure}
    \centering
    \includegraphics[height=8cm,width=\linewidth]{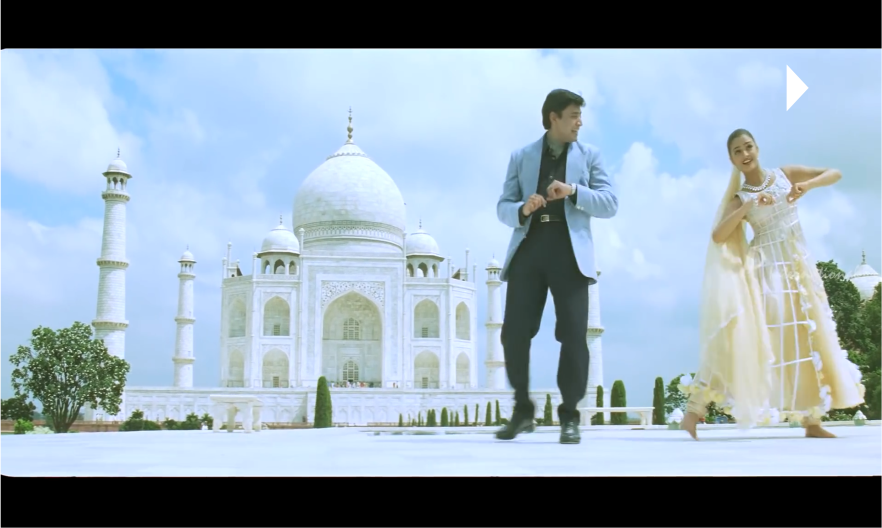}
    \caption{Taj Mahal}
    \label{fig:taj}
\end{figure}

Movies play a pivotal role in influencing tourism, contributing significantly to the appeal of various countries by showcasing famous tourist spots. Many beloved films are set against the backdrop of these tourist destinations, capturing the scenic beauty for memorable dialogue exchanges or song sequences. Filmmakers often seek out renowned tourist spots to lend authenticity and visual splendor to their productions. The knowledge of movies filmed in a location can add an extra layer of excitement to a visit, creating a unique connection between the cinematic world and real-world exploration.

In Figure \ref{fig:lake}, a scene shot at Lake Pichhola, India, is featured in the James Bond series "Octopussy" (1983). Similarly, Figure \ref{fig:taj} showcases a scene filmed at the Taj Mahal, India, from the movie "Jeans" (1998). Our model is designed to accurately detect and identify the specific locations where these scenes were filmed, enriching the viewer's experience by providing informative captions.

Tourism serves as a platform for a country to exhibit its values, history, natural beauty, and wildlife. Movie production becomes a focal point in showcasing these aspects, attracting visitors to explore the real-life settings featured in films. Malta, a European country, recognizes the symbiotic relationship between movies and tourism. Anna Nizio \cite{niziol2009film} highlights that Malta actively supports movie production by offering additional funding of 2\% through the Malta Tourism Authority (MTA) if the produced film is deemed valuable for promoting tourism in Malta. This synergy between the film industry and tourism exemplifies how cinematic creations can become powerful ambassadors for a country's cultural and natural treasures, further solidifying the interplay between movies and the global travel experience. 

The main contributions in this paper are summarized as follows. 

\begin{itemize}
    \item The goal of our proposed model is to enhance the viewer's experience by offering insights into the geographic origins of scenes or songs in movie. 
    \item We contribute a dataset that includes major international tourist destinations, gathered by scraping information from the internet.
\end{itemize}

\section{Related Work}
Changhoon Jeong et al. \cite{jeong2019korean} present the Korean Tourist Spot (KTS) dataset, which is constructed by extracting images from Instagram. This dataset encompasses images along with associated text, hashtags, and likes. The researchers conduct image classification and image captioning using this diverse dataset. Specifically, they use a trained model to generate captions for particular tourist spots based on the classified images. Despite their efforts in image captioning, there remains a limitation for cinema enthusiasts. Individuals passionate about cinema may still encounter challenges in identifying whether their favorite movies were filmed at these tourist spots.

Chi-Seo Jeong et al. \cite{jeong2020deep} propose a deep learning-based recommendation system leveraging social network data to suggest tourist attractions. However, in their system, individuals are required to manually search for similar images from movies they have watched to discover attractions aligned with their interests.

Shanshan Han et al. \cite{han2020extracting} and their team proposed a methodology involving the extraction of relevant images from geotagged photos within tourist destinations. They implemented a filtering process to exclude images unrelated to tourist attractions, such as those taken by locals, among others. The dataset utilized in this research was sourced from Flickr. The researchers combined enhanced cluster methods with multiple deep learning models to achieve their objectives.

Nicole D. Payntar et al. \cite{payntar2021learning} and their team utilized computer vision and machine learning algorithms to analyze tourist movements at archaeological heritage sites in BTC and UNESCO regions, focusing on the aspects of visuality and heritage tourism. Their research aimed to comprehend the extent to which the visuality of tourism sites enhances tourists' expectations. In contrast, this paper takes a different approach by exploring the visual impact movies have on tourism destinations. While Payntar et al. delve into the influence of visuality at archaeological sites, the current study seeks to understand the impact made by movies on the attractiveness and popularity of tourist destinations.

Jaruwan Kanjanasupawan et al. \cite{kan2019prediction} and their team employed a combination of Convolutional Neural Networks (CNN) and Long Short-Term Memory (LSTM) to discern patterns among travelers. They utilized text comments as input data to predict the subsequent tourism spot based on the available data.It is noteworthy that, in contrast to certain methodologies, their model did not consider movie data as a factor in making these predictions. And the model did not require training on previous patterns of traveler behavior.

Movie-induced tourism is a growing trend in which people travel to locations that have been featured in movies. This can be a powerful catalyst for attracting travelers, as the allure of experiencing firsthand the settings depicted in beloved movies can be a strong motivator. The success of such initiatives in various regions emphasizes the symbiotic relationship between the film industry and the tourism sector, highlighting the potential for cinematic storytelling to not only entertain but also to serve as a compelling driver for global travel experiences \cite{evans1997plugging}.

\section{Methodology}
Most feel-good movies incorporate elements like soothing music, picturesque landscapes, or exceptional performances by artists. However, the scenic locations featured in movies are often not explicitly known, unless one possesses prior knowledge. This model, in conjunction with the dataset, serves as the initial step for addressing such scenarios. By utilizing the trained model on the dataset, users can either identify the exact location or find similar locations, offering a valuable tool for those seeking to explore and connect with the beautiful settings depicted in their favorite films. This section consists of the dataset, how it is created, and the convolutional neural network based deep learning model. 

\subsection{Dataset Generation}
We have collected a dataset named MovieTour by extracting images from the internet with a Creative Commons License (CCL). This dataset is collected by using scraping mechanism with the help of Python code over images.The process involves conducting an image search for a specific renowned tourism destination, and the resultant images are extracted from the first page of the search results. Subsequently, a manual curation step is undertaken to segregate unrelated data, ensuring the dataset's relevance and coherence. Following this, the dataset undergoes a transformation process to attain the required shapes, aligning it with the specifications necessary for further analysis and utilization. This approach in dataset collection ensures the accuracy and appropriateness of the data for subsequent applications, such as model training and predictive tasks.

This type of datasets comes handy when watching particular movie, users can employ this model to predict the location of a particular destination or related destinations. The versatility of the model allows it to be applied to movies in any language, provided additional images are incorporated into the dataset to enhance its recognition capabilities. For a comprehensive overview of the tourism destinations recognized by this model, please refer to the accompanying table \ref{tab:dataset}.

\begin{table}[hbt]
    \centering
    \begin{tabular}{|c|c|}
    \hline
    \textbf{Tourist Destinations} & \textbf{Location} \\
    \hline
    Bragatheeswarar Temple & India \\
    Giza Plateau & Egypt \\
    Lake Pichhola & India \\
    Machu Picchu & Peru \\
    Mahabalipuram & India \\
    Marina Beach & India \\
    Meenakshiamman Temple & India \\
    Nilgiri Railway & India \\
    Taj Mahal & India \\
    Pulpit Rock & Norway \\
    Troll Tongue & Norway \\
    Palace of LostCity & South Africa \\
    Petra & Jordan \\
    Leaning Tower of Pisa & Italy \\
    \hline
    \end{tabular}
    \caption{Popular tourist destinations, Name of the country where it is located in the MovieTour dataset.}
    \label{tab:dataset}
\end{table}

From the listed categories, there are a total of 2737 images, with an average of 150 images in each category. This dataset can be utilized for predicting the locations featured in movie scenes.

\subsection{Architecture}
\begin{figure}
    \centering
    \includegraphics[height=7cm,width=\linewidth]{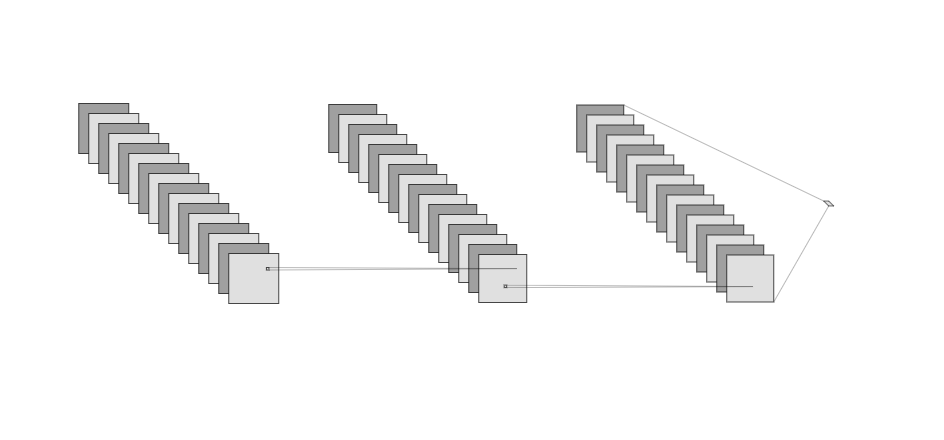}
    \caption{3 Layer CNN}
    \label{fig:enter-label}
\end{figure}

Deep learning, a subset of machine learning, employs multiple layers of nodes to perform non-linear transformations in order to produce output. During training, weights are randomly generated and continually adjusted based on input data. These weights play a crucial role in predicting the outcomes for new, unseen data, classifying them based on knowledge acquired from the training dataset. The training process involves two key phases: forward propagation and backward propagation. In forward propagation, input data and random weights in each node undergo linear calculations, with the results passed to the next layer of nodes after summation. The real magic happens during backward propagation, where the model assesses the resemblance of input data, extracting insights from each training epoch to determine how to process entirely new input data. Figure \ref{fig:enter-label} illustrates the fundamental process of developing a model in deep learning.

In our approach, we employ a three-layer Convolutional Neural Network (CNN) model, with distinct functionalities assigned to each layer. The first layer, serving as both the input layer and the initial stage of feature extraction, consists of 256 nodes. These nodes are designed to detect essential patterns and features within the input data. Moving to the second layer, also comprising 256 nodes, plays a crucial role in further capturing intricate patterns and representations from the features extracted in the initial convolutional layer. Finally, the third layer, with 14 nodes, serves as the output layer consisting of multi layer perceptrons (MLP), responsible for providing the final predictions. This architecture, combining the power of CNN layers for specialized feature extraction and dense layer for intricate pattern recognition, contributes to the model's ability to discern complex relationships within the data, making it well-suited for tasks such as predicting locations of movie scenes in our context.

\section{Conclusion}

By leveraging the CNN model created for training the dataset, we achieved an accuracy of 92.14\%. This success paves the way for further advancements in our research, particularly in the development of a comprehensive model that encompasses all famous tourism destinations. Such an extended model holds immense potential to cater to individuals passionate about travel, providing them with a tool for recognizing and exploring locations worldwide. Additionally, we also suggest that the inclusion of new places in the dataset enhances the model's capability to predict similarly matched places or unseen images within the same destination. This insight underscores the dynamic and adaptable nature of our approach, showcasing its potential utility in diverse scenarios and its contribution to the field of cinematic tourism.

{\small
\bibliographystyle{ieee_fullname}
\bibliography{cvpr}
}

\end{document}